\documentclass[10pt,twocolumn,letterpaper]{article}

\usepackage{iccv}
\usepackage{times}
\usepackage{epsfig}
\usepackage{graphicx}
\usepackage{amsmath}
\usepackage{amssymb}

\usepackage[breaklinks=true,bookmarks=false]{hyperref}

\iccvfinalcopy % *** Uncomment this line for the final submission

\def\iccvPaperID{29} % *** Enter the ICCV Paper ID here

\ificcvfinal\fi

\begin{document}

%%%%%%%%% TITLE
\title{Is There Tradeoff between Spatial and Temporal in Video Super-Resolution?}

\author{Haochen Zhang\quad Dong Liu\thanks{Corresponding author.}\quad Zhiwei Xiong\\
University of Science and Technology of China, Hefei 230027, China\\
{\tt\small zhc12345@mail.ustc.edu.cn, \{dongeliu,zwxiong\}@ustc.edu.cn}
}

\maketitle
\ificcvfinal\thispagestyle{empty}\fi

%%%%%%%%% ABSTRACT
% \begin{abstract}
% Existing video super-resolution (SR) methods are pursuing higher quality of super-resolved video frames, where the quality is usually measured frame by frame in \eg PSNR. These methods often overlook the temporal consistency between video frames. We are interested in analyzing the relationship between two kinds of quality metrics: frame-wise fidelity (called spatial quality) and between-frame consistency (called temporal quality). We benchmark several recent image/video SR methods, using PSNR, SSIM, warping error, and recognition accuracy with a two-stream action recognition network. We also propose a video SR method, where the loss function takes both spatial and temporal metrics into account. Our results imply a tradeoff between spatial and temporal quality in video SR. That says, higher spatial quality may come at the cost of lower temporal quality, to some extent.
% \end{abstract}

%%%%%%%%% BODY TEXT
\section{Motivation}
Recent advances of deep learning lead to great success of image and video super-resolution (SR) methods that are based on convolutional neural networks (CNN). For video SR, advanced algorithms have been proposed to exploit the temporal correlation between low-resolution (LR) video frames, and/or to super-resolve a frame with multiple LR frames \cite{jo2018deep,liu2017robust,liu2018learning,tao2017detail}. These methods pursue higher quality of super-resolved frames, where the quality is usually measured frame by frame in \eg PSNR. However, as mentioned in \cite{lai2018learning}, frame-wise quality may not reveal the consistency between frames. If an algorithm is applied to each frame independently (which is the case of most previous methods), the algorithm may cause temporal inconsistency, which can be observed as flickering. It is a natural requirement to improve both frame-wise fidelity and between-frame consistency, which are termed spatial quality and temporal quality, respectively. Then we may ask, is a method optimized for spatial quality also optimized for temporal quality? Can we optimize the two quality metrics jointly? In short, we want to understand the relationship between spatial quality and temporal quality, in the context of video SR.
%------------------------------------------------------------------------
%------------------------------------------------------------------------
%------------------------------------------------------------------------
%------------------------------------------------------------------------

%\section{Evaluation Regulation}

%------------------------------------------------------------------------
%------------------------------------------------------------------------
%------------------------------------------------------------------------
%------------------------------------------------------------------------
\begin{figure*}
   \centering
   \includegraphics[width=\linewidth]{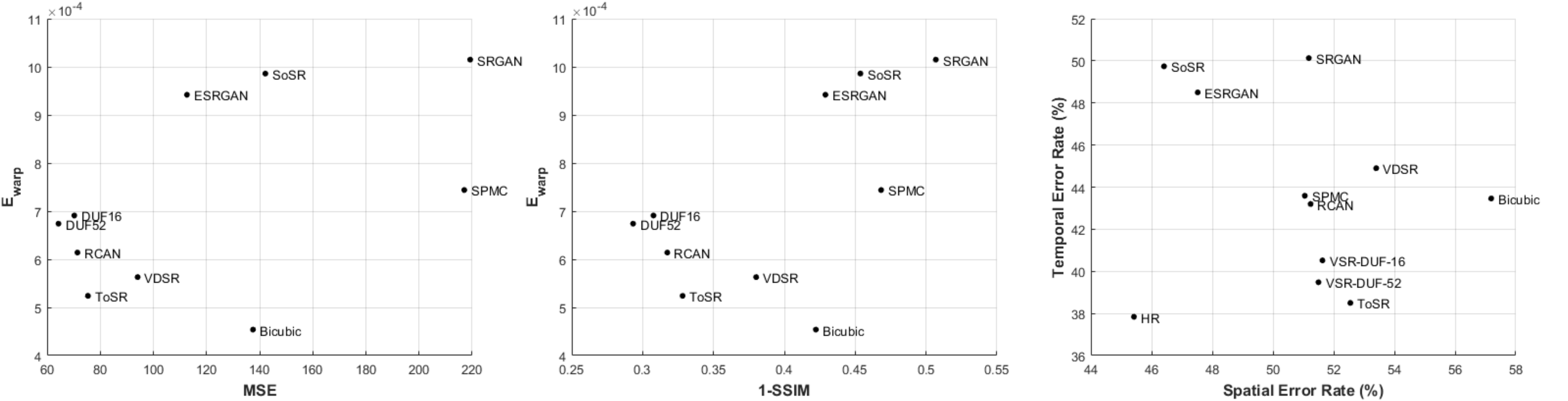}
   \caption{Evaluation results with different spatial and temporal quality metrics, including MSE, 1$-$SSIM, warping error, recognition error rate in spatial/temporal stream. In each plot, the bottomleft corner is the best. In the third plot, error rates of HR videos are shown for reference.}
   \label{exp1}
\end{figure*}
\begin{figure*}
   \centering
   \includegraphics[width=\linewidth]{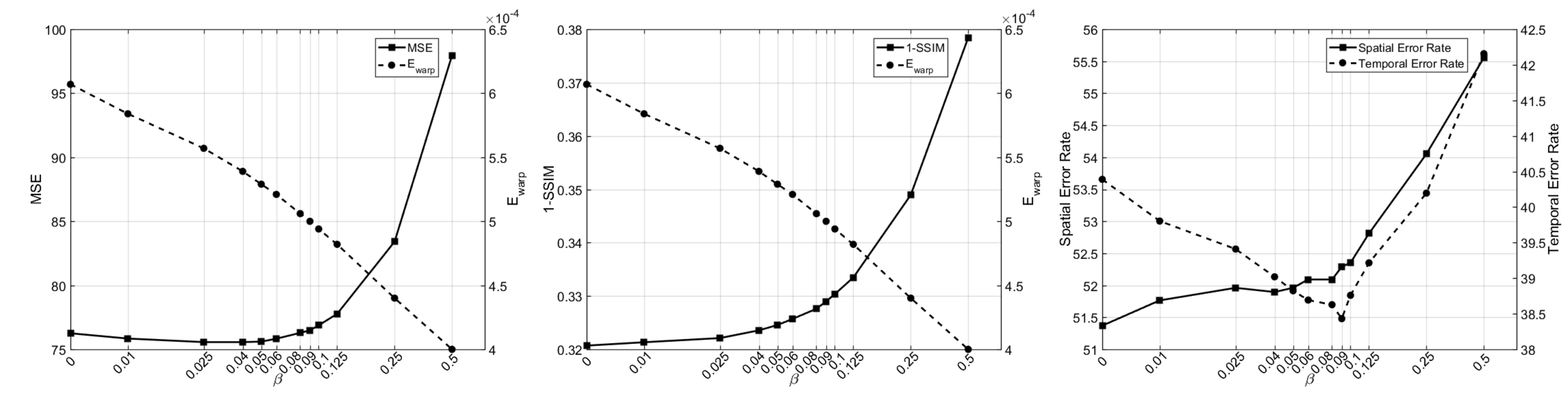}
   \caption{Performance of our SR method trained with (\ref{warping_loss}) and different $\beta$ values. The horizontal axis is shown in logarithmic scale.}
   \label{exp2}
\end{figure*}

\section{Experiments and Analyses}
\textbf{Dataset.} We use the HMDB51 dataset, which is a collection of real-world videos and was widely used for action recognition research \cite{kuehne2011hmdb}. The dataset includes 6,766 video clips that belong to 51 action categories. The dataset provides three training/testing splits. We use split1 as a representative. We down-sample the video clips by a factor of 4 using bicubic interpolation, and then super-resolve the video clips with different methods.

\textbf{Compared methods.} As a naive baseline we use bicubic interpolation to up-sample. We test four image SR methods: VDSR \cite{kim2016accurate}, RCAN \cite{zhang2018image}, SRGAN \cite{ledig2017photo}, and ESRGAN \cite{wang2018esrgan}, where we super-resolve each frame independently. We test four video SR methods: SPMC \cite{tao2017detail}, DUF \cite{jo2018deep}, SoSR, and ToSR. SoSR and ToSR are proposed by ourselves as new video SR methods for facilitating action recognition rather than for PSNR \cite{zhang2019two}. All the compared methods, excluding bicubic, are based on CNN. For each method we use the pretrained model provided by the corresponding authors. It is worth noting that the compared methods use different training data, but none of the training data has overlap with HMDB51.

\textbf{Spatial quality metrics.} For spatial quality, we consider MSE and SSIM. They are calculated by comparing the super-resolved videos against the original videos frame by frame. For each frame, MSE and SSIM are calculated on RGB and luma component, respectively. Then, they are averaged over each video and then over the entire test set. Moreover, we consider the quality not only at the signal level, but also at the semantic level. We use a pretrained action recognition network, TSN \cite{wang2016temporal}, to evaluate the action recognition accuracy based on the super-resolved videos. Our used TSN model is trained with HMDB51 split1 training set. TSN is a two-stream network, so for spatial quality we use its spatial stream, \ie action recognition based on the frames.

\textbf{Temporal quality metrics.} For temporal quality, we use the warping error proposed in \cite{lai2018learning}, which is calculated for a super-resolved video $V$: 
\begin{equation}
\label{warping_error}
\begin{aligned}
&E_{warp}(V)=\frac{1}{T-1}\sum_{t=1}^{T-1}\frac{1}{\sum_{p=1}^{N}M_t(p)}E_{warp}(V_t,V_{t+1}) \\
&E_{warp}(V_t,V_{t+1})=\sum_{p=1}^{N}M_t(p)[V_{t}(p)-V_{t+1}^w(p)]^2
\end{aligned}
\end{equation}
where $T$ is the number of frames. $V_{t}$ and $V_{t+1}$ are two consecutive frames in the video. $N$ is the number of pixels per frame, and $p$ denotes each pixel. $M_t$ is a mask standing for whether each pixel is occluded or not. $V_{t+1}^w$ is a \emph{warped} frame, \ie $V_{t+1}^w(p)=V_{t+1}(p+F_{t\rightarrow t+1}(p))$, where $F_{t\rightarrow t+1}$ is the optical flow from $V_{t}$ to $V_{t+1}$. Optical flow is calculated by Flownet2.0 \cite{ilg2017flownet}. Occlusion mask is estimated by \cite{ruder2016artistic}. In addition, we use the temporal stream of TSN \cite{wang2016temporal} to evaluate the action recognition accuracy based on the optical flows extracted from the super-resolved videos.

\textbf{Results.} Figure \ref{exp1} displays the evaluation results of the spatial and temporal quality metrics for each method.
For spatial quality, the relative trends of MSE and SSIM are almost consistent, so we analyze the MSE.
Of these methods, DUF, RCAN, and VDSR are optimized for MSE.
We can find that in terms of MSE, DUF is the best and VDSR is the worst among the three, and all the three are far better than bicubic.
However, in terms of warping error, DUF is the worst and VDSR is the best among the three.
Indeed, bicubic performs the best among all the compared methods in terms of warping error!
This can be understood, since bicubic tends to generate oversmooth frames, which have less temporal inconsistency. To improve spatial quality, advanced SR methods try to add details into the frames, but take the risk of producing flickering artifacts.
This seems to indicate a tradeoff between spatial and temporal quality.
Moreover, we look at the spatial/temporal quality metrics evaluated by recognition accuracy. The best methods in the spatial stream, SoSR and ESRGAN, all use adversarial loss in their training. However, these methods perform the worst in the temporal stream. These methods also lead to very high warping error. This seems another evidence of the spatial-temporal tradeoff.
It is worth noting that the tradeoff is only to some extent. DUF with 52 layers is consistently better than DUF with 16 layers, ESRGAN is consistently better than SRGAN, in every considered metric. The consistently better performance is achieved at the cost of much increased network complexity.
%------------------------------------------------------------------------

\textbf{Joint Optimization of Spatial and Temporal.}
We extend the ToSR method, where we use a siamese network to super-resolve two consecutive frames simultaneously. We use the following loss function:
\begin{equation}
% \begin{small}
\label{warping_loss}
\mathcal{L}=\alpha \|\hat{I}_{t}-I_{t}\|_F^2+\alpha \|\hat{I}_{t+1}-I_{t+1}\|_F^2+\beta E_{warp}(\hat{I}_t,\hat{I}_{t+1})
% \end{small}
\end{equation}
where $I$ and $\hat{I}$ denote HR and SR frames. $E_{warp}(\hat{I}_t,\hat{I}_{t+1})$ is similar to that defined in (\ref{warping_error}), except that the mask is defined as $M_t(p)=\exp(-50[I_{t}(p)-I_{t+1}^w(p)]^2)$ \cite{lai2018learning}.
We conduct experiments with fixed $\alpha=0.5$ and variable $\beta$.
The results are shown in Figure \ref{exp2}.
As $\beta$ increases, warping error decreases monotonously, but MSE or 1$-$SSIM increases; error rate in the spatial stream increases, but error rate in the temporal stream decreases to some extent.
In summary, it seems a difficulty to optimize the spatial and temporal quality metrics simultaneously.
%------------------------------------------------------------------------
%------------------------------------------------------------------------
%------------------------------------------------------------------------
%------------------------------------------------------------------------

\section{Conclusion}
In \cite{blau2018perception}, it was proved that minimizing distortion and optimizing perceptual naturalness can be contradictory, which was named perception-distortion tradeoff.
In \cite{liu2019classification}, the tradeoff is extended to classification-distortion-perception. Similarly, our empirical results imply a tradeoff between spatial and temporal in video SR. We are seeking a theoretical proof.
%------------------------------------------------------------------------
%------------------------------------------------------------------------
%------------------------------------------------------------------------

\end{document}